\documentclass[a4paper, 10pt, conference]{IEEEtran}
\IEEEoverridecommandlockouts
\usepackage{cite}
\usepackage{url}
\usepackage{amsmath,amssymb,amsfonts}
\usepackage{algorithmic}
\usepackage{graphicx}
\usepackage{textcomp}
\usepackage{makecell}
\usepackage{multirow}
\usepackage{hyperref}
\usepackage{xcolor}
\def\BibTeX{{\rm B\kern-.05em{\sc i\kern-.025em b}\kern-.08em
    T\kern-.1667em\lower.7ex\hbox{E}\kern-.125emX}}

\usepackage{lipsum}

\newcommand\blfootnote[1]{%
  \begingroup
  \renewcommand\thefootnote{}\footnote{#1}%
  \addtocounter{footnote}{-1}%
  \endgroup
}

\begin{document}

\title{Towards Graph Representation Learning Based Surgical Workflow Anticipation}
\author{
\IEEEauthorblockN{Xiatian Zhang$^{1}$, Noura Al Moubayed$^{1}$, Hubert P. H. Shum$^{1,\dag}$}
\IEEEauthorblockA{\small $^1$ Department of Computer Sciences, Durham University, Durham, United Kingdom}
\vspace{-2.0em}
}


\maketitle
\begin{abstract}
Surgical workflow anticipation can give predictions on what steps to conduct or what instruments to use next, which is an essential part of the computer-assisted intervention system for surgery, e.g. workflow reasoning in robotic surgery. However, current approaches are limited to their insufficient expressive power for relationships between instruments. Hence, we propose a graph representation learning framework to comprehensively represent instrument motions in the surgical workflow anticipation problem. In our proposed graph representation, we maps the bounding box information of instruments to the graph nodes in the consecutive frames and build inter-frame/inter-instrument graph edges to represent the trajectory and interaction of the instruments over time. This design enhances the ability of our network on modeling both the spatial and temporal patterns of surgical instruments and their interactions. In addition, we design a multi-horizon learning strategy to balance the understanding of various horizons indifferent anticipation tasks, which significantly improves the model performance in anticipation with various horizons. Experiments on the Cholec80 dataset demonstrate the performance of our proposed method can exceed the state-of-the-art method based on richer backbones, especially in instrument anticipation (1.27 v.s. 1.48 for inMAE; 1.48 v.s. 2.68 for eMAE). To the best of our knowledge, we are the first to introduce a spatial-temporal graph representation into surgical workflow anticipation.
\end{abstract}
\begin{IEEEkeywords}
Surgical Data Science, Surgical Workflow Analysis, Deep Learning, Graph Representation Learning
\vspace{-1.6em}
\end{IEEEkeywords}
\blfootnote{Emails: \{xiatian.zhang, noura.al-moubayed, hubert.shum\}@durham.ac.uk}
\blfootnote{$^\dag$ Corresponding author}

\section{Introduction}
Surgical workflow anticipation is a highly essential task for surgical workflow analysis in the computer-assisted intervention (CAI) systems for surgery. Firstly, anticipating the occurrence of surgery instruments and phase can be regarded as a kind of \emph{Context-aware Assistance} in surgery \cite{maier2017surgical}, which can improve the performance of surgery teams. For instance, the anticipation for instrument occurrence can support a more efficient collaboration in the surgery team. Secondly, the surgical workflow anticipation is also a key component for automatic surgery intervention, which can bridge the understanding between current and future scenarios. That can eventually facilitate automatic surgery decision-making like the intervention time selection in the CAI system \cite{rivoir2020rethinking}.

Currently, various works have investigated anticipation of surgical workflow \cite{rivoir2020rethinking,twinanda2018rsdnet,bodenstedt2019prediction,rivoir2019unsupervised,ban2021aggregating,yuan2021surgical,marafioti2021catanet}. Most works are solely based on pixel-level visual features extracted by ResNet \cite{he2016deep} and similar backbones, and learn these features directly with temporal models \cite{schmidhuber1997long}. This design may ignore semantic information like the interaction between instruments \cite{yuan2021surgical}, which gives a rise to the limitation in expressing high-level scenario information like the surgeon’s intention. 

The state-of-the-art in surgical workflow anticipation for instruments and phases, IIA-Net \cite{yuan2021surgical},  used the non-visual features. It proposed an instrument interaction module into their feature extractor to reflect instrument interactions via geometric relations of instrument bounding boxes and semantic segmentation maps. However, this design ignores the interaction between different instruments and cannot represent complex relationships in the real scenario. Additionally, IIA-Net \cite{yuan2021surgical} still use visual features and also introduce the ground-truth annotation into their features for temporal modeling via Multi-Stage Temporal Convolutional Network (MSTCN) \cite{farha2019ms}. That lets this work need more backbones for features learning and phase/tool recognition when deployed in an automatic CAI system, which may limit their actual performance in real-world scenarios.

Recent works also attempted to introduce non-visual features into surgery activity analysis via graph representation and graph neural networks (GNNs) \cite{islam2020learning,sarikaya2020towards}, due to highly expressive power for complex abstract relationships in graphs. The data structure in graph represents a set of objects (nodes) and their relationships (edges) \cite{zhou2020graph},  and GNNs are the widely modeling method for the graph data due to their demonstrated performance \cite{zhou2020graph}. \cite{islam2020learning} represents the interactions between the defective tissue and surgical instruments in a form of the graph structure, and learn the graph data with inductive graph representation learning \cite{hamilton2017inductive}. \cite{sarikaya2020towards} represents the joint pose estimations of surgical tools into a spatial-temporal graph, and model with graph convolutions \cite{yan2018spatial}, which can learn the spatial and temporal patterns from the graph nodes and their neighbors. These previous attempts demonstrate graph structure can contain enough information for surgical activity understanding. 

In our works, we hypothesize that the graph structure of surgical instruments can reason the future surgical activity like instrument presence or surgery phase in surgical workflow anticipation.  To express a more complex relationship between surgery instruments than \cite{yuan2021surgical}, we set the instrument interactions as the inter-node edges in our spatial-temporal graph structure for surgery instruments. Considering both spatial and temporal features are critical for surgical workflow anticipation, our works implement the graph convolution \cite{yan2018spatial} and MSTCN \cite{farha2019ms} into modeling this graph representation. Our approach is just based on graph structure data of instrument bounding box information for data-efficient learning. To the best of our knowledge, we are the first to introduce graph representation learning into surgical workflow anticipation.

\begin{figure*}[h]
\centering
\includegraphics[scale = 0.35]{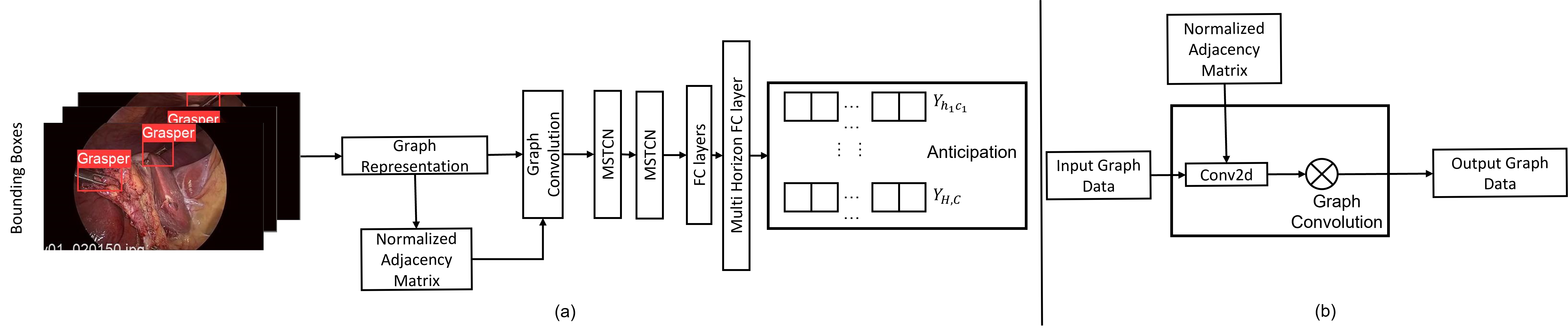}
\caption{(a) Network Overview; (b) The Graph Convolution}
\label{fig:Net}

\end{figure*}

Our source code is available on {\color{blue}\url{https://github.com/summerwings/GCN_Anticipation}}. The contributions of this work can be summarized as follows:
\begin{enumerate}
  \item We propose the spatial-temporal graph representation into surgical workflow anticipation, which can represent the interaction information of surgery instruments better than previous approaches.
  \item We combine the graph convolution and MSTCN to implement a network that anticipate the occurrence of surgical instruments/phases from graph structure data, of which the performance exceeds or can be comparable to the state-of-the-art approach.
  \item We design a multi horizon learning strategy to balance the understanding of various horizons in different anticipation tasks, which significantly improves the graph convolutional network performance in anticipation with various horizons.
\end{enumerate}


\section{Methodology}
\subsection{Task Formulation}
The anticipation task can be formulated as a regression task \cite{rivoir2020rethinking,yuan2021surgical}. Given a frame $t$ from video $x$, we firstly extract instrument bounding boxes $b_{t}$. Given the observed sequence $\left\{ (x_{t}, b_{t}) \cdots (x_{T_{obs}}, b_{T_{obs}}) \right\}$ from time $0$ to time $T_{obs}$, our model predicts the remaining time until the occurrence of instrument $\tau$ and phase $p$. The ground truth $r(x_{T_{obs}}, \tau \slash p)$ ranges $\left[0,h\right]$, where 0 denotes $\tau \slash p$ is happening and $h$ denotes the \textbf{horizon} that $\tau \slash p$ would not happen within $h$ minutes. 

\subsection{Network Architecture}
Figure \ref{fig:Net} (a) shows an overview of our network. It firstly inputs the bounding boxes of instruments detected by object detection backbones (e.g. YOLOv5  \cite{glenn_jocher_2022_6222936}) of each video sequence, and represent the bounding boxes into a spatial-temporal graph. Then, this network uses the graph convolution \cite{yan2018spatial} and the MSTCN \cite{farha2019ms} to model a spatial-temporal graph representation. Finally, it uses a multi-horizon fully connected (FC) layer to regress the anticipation for instruments or phases for each frame in the input sequence.

\begin{figure}[h]
\begin{center}
\includegraphics[scale = 0.40]{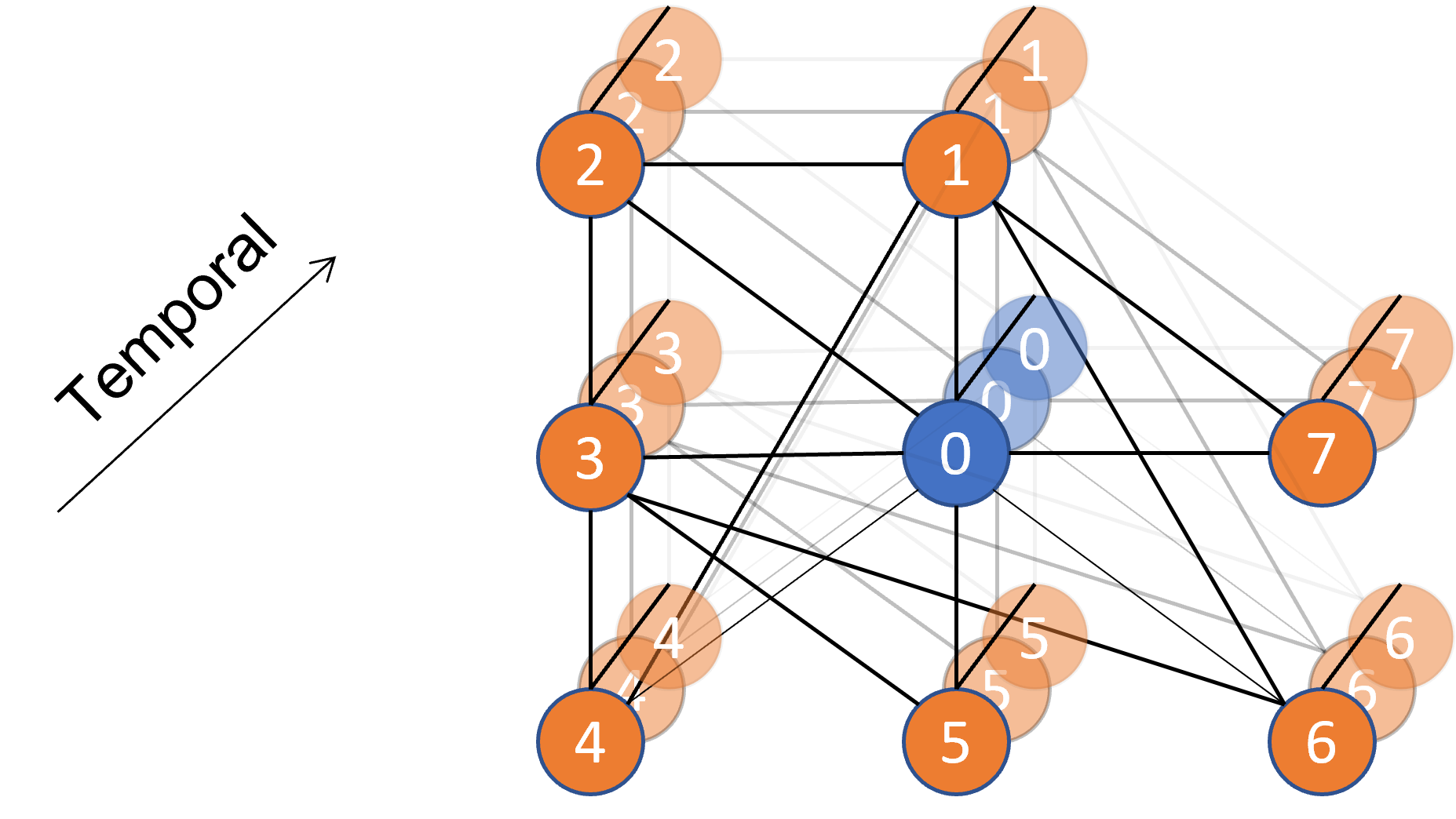}
\caption{Graph Representation 0: Center Viewpoint; 1: Grasper; 2: Bipolar; 3: Hook; 4: Scissors; 5: Clipper; 6: Irrigator; 7: SpecimenBag.}
\label{fig:graph_c}
\end{center}
\end{figure}
\subsubsection{The Graph Representation Strategy}
In our graph representation for surgery scenario, the complete graph structure of instruments are modeled in the form of $G(V,E)$. The node set $V = \left\{ v_{ti}| t = 1, \cdots T_{obs} , i = 1, \cdots , j, \cdots , N \right\}$  includes the all the nodes $i$ in an observed video sequence, where $N$ denotes the total number of defined nodes in graph. Besides the instrument bounding boxes, our proposed graph also includes the center viewpoint of videos as one node to provide a fixed surrounding reference for the instrument nodes and represent the interaction between instruments and the surrounding scene. The feature vector on a node is:
\begin{equation}
F_{v_{ti}} = (c_{x_{t}}, c_{y_{t}}, w_{t}, h_{t}),
\end{equation}
where $c_{x_{t}}$ and $c_{y_{t}}$ respectively denote the coordinates of center point $x/y$ in in frame $t$, $w_{t}$ and $h_{t}$ respectively denote the detected width and height of each tool in frame $t$. 

The edge set $E$ consists of two subsets. The first edge set represents the inter-node edges at each frame, denoted as $E_{v} = \left\{ v_{ti}v_{tj}| i, j \in H, i \neq j \right\}$, where $H$ is the set of defined interaction between nodes (instruments or the center viewpoint). These inter-node edges represent the interaction between the center viewpoint/the constantly existing instrument defined by prior medical knowledge (e.g. grasper in the cholecystectomy surgeries) and other instruments.  These inter-node connections in each frame can be represented in a $N \times N$ \emph{Adjacency Matrix} $A$ where $A^{ij} = 1$ if $ v_{ti}v_{tj} \in E_{v}$. The second edge set represents the inter-frame edges for each node, denoted as $E_{F} = \left\{ v_{ti}v_{(t+1)i} \right\}$. These inter-frame edges represent the temporal dependency of instrument motions. Figure \ref{fig:graph_c} shows an example of our graph representation for laparoscopic cholecystectomy from Cholec80 dataset \cite{twinanda2016single}. We consider the interaction between center viewpoint, graspers, hooks, and other instruments as the inter-node edges in our proposed graph structure, due to their almost always presence of graspers and hooks across surgery and their constant interaction with other tools \cite{yuan2021surgical}.

\subsubsection{Graph Convolution}
Figure \ref{fig:Net} (b) shows the structure of the graph convolution (GC) modified from \cite{yan2018spatial} in our network to learn  inter-node relationships at each frame. The input graph data $F_{in}$ in GC is processed with the following formula to obtain the output graph data $F_{out}$ \cite{kipf2017semi}:
\begin{equation}
F_{out} = \Lambda^{-1/2}( A + I) \Lambda^{-1/2} F_{in}W,
\end{equation}
where $\Lambda^{ii} = \Sigma_{j}A^{ij} + I^{ij}$, $A$ is the adjacency matrix, $I$ is an identity matrix and $\Lambda^{-1/2}( A + I) \Lambda^{-1/2}$ is the normalized adjacency matrix, $W$ is the matrix including the multi-channel weight vectors for calculating output graph data.

\subsubsection{Multi-Stage Temporal Convolutional Network}
The temporal inter-frame relationships of GC output graph data will be modeled by MSTCN. This includes multi-stage multi-layer temporal convolutional networks with dilated convolutions \cite{farha2019ms}. In our network, we use the causal dilated convolution in MSTCN to ensure the online inference in anticipation tasks  \cite{yuan2021surgical}, where anticipation in each frame only relies on the previous and current information. Compared with IIA-Net \cite{yuan2021surgical}, we still set two stages for MSTCN but set the layer number of each stage as 14 instead of 10, to ensure our network can summarize longer temporal information for different video sequences of various lengths.

\subsubsection{Multi Horizon Learning}
Surgical workflow anticipation tasks normally contain the short-horizon tasks and long-horizon tasks. Previous works train these tasks with the MAE (mean absolute error) of the anticipated remain time v.s true remain time directly or the remaining time based on the ground truth of max-horizon anticipation. Hence, the optimization for the relatively small errors in specified horizon anticipations can be disturbed by the large errors in other horizon anticipations. Hence, to balance the understanding of various horizons in different anticipation tasks, we propose a multi horizon learning strategy. We use a multi horizon FC layer to output our final results in different horizon levels. In this FC layer, the output of one frame $t$ is in the following form:
\begin{equation}
Y_{t} = (Y_{h_{1}, c_{1}}; Y_{h_{1}, c_{2}} ;\cdots ;Y_{h_{H}, c_{C-1}}; Y_{h_{H}, c_{C}}) \in \mathbb{R}^{H \times C},
\end{equation}
where $h \in H$ is the set of horizons and $c \in C$ is the set of instruments or phases to anticipate. An example is the horizon FC layer setting in our experiment for Cholec80 dataset \cite{twinanda2016single}. In our experiment, we anticipate the future occurrence of instruments or phases with three horizons of 2, 3, and 5 minutes. Then, our multi-horizon FC layer outputs the vector with the anticipation results of three horizons.

As the anticipation needs to consider performance in both occurrence and non-occurrence in short-term and long-term horizons, the anticipation loss would use the combination of $wMAE$ \cite{yuan2021surgical} which is the mean MAE of where ground truth ranges $(0,h)$ and where ground truth is out of $h$, $inMAE$ \cite{yuan2021surgical} where ground truth ranges $(0,h)$, $inMAE$ \cite{yuan2021surgical} where ground truth ranges $(0,h)$, $pMAE$ \cite{rivoir2020rethinking} where predicted anticipation ranges $(0.1h,0.9h)$ and $eMAE$ \cite{yuan2021surgical} where ground truth ranges $(0,0.1h)$. In optimization, different horizon outputs are back propagated according to loss with corresponding horizons. Hence, the loss in training would be: 
\begin{equation}
Loss = \Sigma \alpha wMAE_{h} + \beta inMAE_{h} + \gamma pMAE_{h} + \delta eMAE_{h},
\end{equation}
where $h$ denotes the corresponding horizon, and $\alpha, \beta, \gamma, \delta$ denote the corresponding weight for each loss.

\section{Experiment and Results}
\subsection{Experiment Setup}
We evaluated our method on the publicly available  Cholec80 dataset (\url{http://camma.u-strasbg.fr/datasets}, \cite{twinanda2016single}), which includes laparoscopic cholecystectomy procedures. It consists of 80 videos ranging from 15 min to 90 min. We separated the dataset to 40/20/20 for training, validation, and testing. We resampled the video from 25 fps to 1 fps and inputted the whole video for backward propagation. To detect the surgical instrument bounding boxes, we trained a YOLOv5 object detector \cite{glenn_jocher_2022_6222936} from the dataset m2cai16-tool-locations (\url{https://ai.stanford.edu/~syyeung/resources/m2cai16-tool-locations.zip}, \cite{jin2018tool}). The graph representation in experiments is as Figure \ref{fig:graph_c}. In training, the Adam optimizer was used for training.  The hyper-parameters epoch, learning rate, and batch size were set as 100, 0.002, and 1, respectively. The training loss weights $\alpha, \beta, \gamma, \delta$ were set as 0.9, 0.1, 0.8 and 0.3, respectively. The hyper-paramaters were selected by Bayesian hyper-parameter search method via the Weights \& Biases platform \cite{wandb}. We implemented our method with PyTorch 1.10.1 and trained models using one Nvidia GeForce GTX 2080 Ti GPU.

The horizon set is $\left\{2, 3, 5 \right\}$ in our experiment, to let our method be compared with previous works \cite{yuan2021surgical}. Referring the state-of-the-art study \cite{yuan2021surgical}, we evaluate the performance of our method by $inMAE$, $pMAE$ and $eMAE$ for $h \in \left\{2, 3, 5 \right\}$. All evaluations were based on the online inference mode, which means anticipation in each frame can only rely on the previous and current information. Our anticipation targets will not include graspers, hooks due to their nearly constant presence, and the preparation phase due to its always first position.
\subsection{Results and Discussion}
\begin{table}[h]
    \centering
        \caption{Surgery instrument anticipation comparison of $inMAE$/$pMAE$/$eMAE$}
\begin{tabular}{c|c|c|c}
\hline
\multirow{2}{*}{ }&\multicolumn{3}{c}{$inMAE$/$pMAE$/$eMAE$}\\
\cline{2-4}
 &2 min & 3 min & 5 min \\
\hline
\makecell{Bayesian \\CNN-LSTM\cite{rivoir2020rethinking,yuan2021surgical}} &0.77/0.64/1.12 & 1.13/0.92/1.65& 1.80/1.49/2.68\\
\hline
IIA-Net \cite{yuan2021surgical} &0.66/\textbf{0.42}/1.01 & 0.97/\textbf{0.69}/1.46 & 1.48/\textbf{1.28}/2.14 \\
\hline
Ours &\textbf{0.57}/0.47/\textbf{0.65} & \textbf{0.81}/0.74/\textbf{0.94} & \textbf{1.27}/1.32/\textbf{1.48} \\
\hline
    \end{tabular}
    \label{tab:exp}
\end{table}
\begin{table}[h]
    \centering
        \caption{Surgery Phase anticipation comparison of $inMAE$/$pMAE$/$eMAE$}
\begin{tabular}{c|c|c|c}
\hline
\multirow{2}{*}{ }&\multicolumn{3}{c}{$inMAE$/$pMAE$/$eMAE$}\\
\cline{2-4}
 &2 min & 3 min & 5 min \\
\hline
\makecell{Bayesian \\CNN-LSTM\cite{rivoir2020rethinking,yuan2021surgical}} &0.63/0.62/1.02 & 0.86/0.85/1.47 & 1.17/1.37/1.54 \\
\hline
IIA-Net \cite{yuan2021surgical}&  0.62/0.49/1.18& 0.81/0.73/1.42& \textbf{1.08}/1.22/1.09\\
\hline
Ours &\textbf{0.54}/\textbf{0.47}/\textbf{0.47} & \textbf{0.75}/\textbf{0.70}/\textbf{0.64} & 1.11/\textbf{1.21}/\textbf{0.90} \\
\hline
    \end{tabular}
    \label{tab:exp1}
\end{table}
\begin{table*}[h]
    \centering
        \caption{Ablation experiments for our method. GC: Graph Convolution, $GPK$: Graph representation with prior knowledge, TC: MSTCN, HL$_{h=2 min}$, HL$_{h=3 min}$, HL$_{h=5 min}$: Horizon Learning for $h$ = 2 min, 3 min, and 5 min, respectively.}
        
\begin{tabular}{c|c|c|c|c|c|c|c|c}
\hline
\multicolumn{6}{c|}{}&\multicolumn{3}{c}{Instrument  Anticipation ($inMAE$)}\\
\cline{1-5}
\hline
GC& GPK&TC&HL$_{h=2 min}$&HL$_{h=3 min}$&HL$_{h=5 min}$& h = 2 min& h = 3 min & h =5 min\\

\hline
\checkmark & &  &&&  & 0.90 & 1.33 & 2.26\\
\hline
\checkmark &\checkmark&&&& &0.89 & 1.33 & 2.27\\
\hline
 &&\checkmark&&& &0.96 & 1.30 & 2.55\\
\hline
\checkmark &&\checkmark&&& &0.90 & 1.42 & 1.97\\ 
\hline
\checkmark &\checkmark&\checkmark&&& &0.80 & 1.26 & 2.21\\
\hline
\checkmark&\checkmark &\checkmark&\checkmark&& &0.59 & 1.35 & 2.18\\
\hline
\checkmark &\checkmark&\checkmark&&\checkmark& &0.91 & 0.83 & 2.26\\
\hline
\checkmark&\checkmark &\checkmark&&&\checkmark &0.87 & 1.33 & 1.32\\
\hline
\checkmark &\checkmark&\checkmark&\checkmark&\checkmark&\checkmark &\textbf{0.57} &\textbf{0.81} &\textbf{1.27}\\

\hline
    \end{tabular}
    \label{tab:ab_study}
\end{table*}

\subsubsection{Comparison with State-of-the-Art}
Table \ref{tab:exp} and Table \ref{tab:exp1} shows the overview results of comparison with the state-of-the-art. In instrument anticipation, our method can anticipate better in $inMAE/eMAE$ or be comparable with the state-of-the-art method in $pMAE$ (The difference is not larger than 0.05). In phase anticipation, our method can generally
anticipate better in all metrics in short-term anticipation and be also comparable in long-term anticipation (The difference in $inMAE$ is not larger than 0.05). In summary, these results indicate that our current graph representation for surgery scenarios via bounding boxes can effectively represent the future status of surgery scenarios, especially when considering our model use fewer backbones and less ground-truth information than the state-of-the-art method.
\subsubsection{Running Time Test}
On the GTX 2080 Ti GPU we used, the mean over inference time of our model for each frame is within 0.030s, which indicates the real-time performance of our method in normal consume computers.
\subsubsection{Ablation Study}
We designed an ablation study of our method to explore the effect of different components (The fully-connected graph representation was used when ablating the GPK). Table \ref{tab:ab_study} shows these components are complementary and our proposed model with all components performs effective anticipation in both short/long-term horizons. It also highlights our horizon learning strategy can improve the model performance in anticipation for various horizons.
\section{Conclusion}
In this paper, we propose a graph representation learning based way for surgical workflow anticipation. It shows that the graph representation is effective to resolve surgical instrument anticipation. Without rich backbones, our model is a strong baseline for the surgical anticipation works, especially in short horizons. Furthermore, our multi-horizon learning schema provides a solution to perform anticipation tasks within different horizons. As a lightweight model, our model can be transferred to real clinical scenarios at a low cost and provide a more smooth collaboration among the surgery team to improve the surgery performance. Future work includes a more dynamic representation of surrounding surgical information to improve long-horizon anticipations.



\bibliographystyle{IEEEtran}
\bibliography{mybib}

\end{document}